%% file: anonymous-submission-latex-2026.tex
\documentclass[letterpaper]{article} 
\usepackage[draft]{aaai2026}  
\usepackage{newtxtext,newtxmath} 
\usepackage{helvet}  
\usepackage{courier}  
\usepackage[hyphens]{url}  
\usepackage{graphicx} 
\usepackage{natbib}  
\usepackage{caption} 
\usepackage{amsmath}

\usepackage{amssymb}
\usepackage{booktabs}
\usepackage{multirow}
\usepackage[table]{xcolor}
\definecolor{lightgray}{gray}{0.9}
\usepackage{algorithm}
\usepackage{algorithmic}

\usepackage{newfloat}
\usepackage{listings}
\DeclareCaptionStyle{ruled}{labelfont=normalfont,labelsep=colon,strut=off} 
\floatstyle{ruled}
\newfloat{listing}{tb}{lst}{}
\floatname{listing}{Listing}
\title{Make LVLMs Focus: Context-Aware Attention Modulation for Better Multimodal In-Context Learning}
\author{
    Yanshu Li\textsuperscript{\rm 1},
    Jianjiang Yang\textsuperscript{\rm 2},
    Ziteng Yang\textsuperscript{\rm 1},
    Bozheng Li\textsuperscript{\rm 1},
    Ligong Han\textsuperscript{\rm 3},
    Hongyang He\textsuperscript{\rm 4},
    Zhengtao Yao\textsuperscript{\rm 5},
    Yingjie Victor Chen\textsuperscript{\rm 6},
    Songlin Fei\textsuperscript{\rm 6},
    Dongfang Liu\textsuperscript{\rm 7},
    Ruixiang Tang\textsuperscript{\rm 8}\thanks{Corresponding Author.}
}
\affiliations{
    \textsuperscript{\rm 1}Brown University\\
    \textsuperscript{\rm 2}University of Bristol\\
    \textsuperscript{\rm 3}MIT-IBM Watson AI Lab\\
    \textsuperscript{\rm 4}University of Warwick\\
    \textsuperscript{\rm 5}University of Southern California\\
    \textsuperscript{\rm 6}Purdue University\\
    \textsuperscript{\rm 7}Rochester Institute of Technology\\
    \textsuperscript{\rm 8}Rutgers University\\
    yanshu\_li1@brown.edu, ruixiang.tang@rutgers.edu
    
}

\usepackage{bibentry}

\begin{document}

\maketitle

\begin{abstract}
Multimodal in-context learning (ICL) is becoming a key capability that allows large vision-language models (LVLMs) to adapt to novel tasks without parameter updates, which expands their usefulness in many real-world applications. However, ICL performance remains unstable even when the in-context demonstrations (ICDs) are well matched, showing that LVLMs still struggle to make full use of the provided context. While existing work mainly focuses on prompt engineering or post-hoc logit calibration, we study the attention mechanisms inside LVLMs to address their inherent limitations. We identify two important weaknesses in their self-attention that hinder effective ICL. To address these weaknesses, we propose \textbf{Context-Aware Modulated Attention} (CAMA), a training-free and plug-and-play method that dynamically adjusts attention logits based on the input in-context sequence. CAMA uses a two-stage modulation process that strengthens attention to semantically important tokens, especially visual ones. Across four LVLMs and seven benchmarks, CAMA consistently outperforms vanilla models and baselines, showing clear effectiveness and generalization. It can also activate the intended benefits of prompt engineering methods and remains robust across different sequence configurations. Therefore, CAMA opens up new directions for improving multimodal reasoning through a deeper understanding of attention dynamics.
\end{abstract}

\section{Introduction}

\label{sec:intro}
Large vision-language models (LVLMs) have emerged as powerful tools for multimodal information processing and generation \cite{LVLM}. Through large-scale pretraining, they integrate visual and textual signals into the shared representation space of large language models (LLMs) and have achieved notable success across vision-language tasks \cite{lvlm1,lvlm2}. However, adapting LVLMs to new domains remains challenging due to the high costs of multimodal data preparation and training.

To mitigate these costs, researchers are applying in-context learning (ICL), a technique widely used in LLMs \cite{iclr1,iclr2}, to LVLMs \cite{icl3}. In ICL, a few in-context demonstrations (ICDs) are incorporated into the input as reference examples, enabling the model to adapt to new tasks by interpreting these demonstrations without parameter updates. Recent advancements in model architectures and training protocols have enabled LVLMs to process multiple images and perform interleaved reasoning, making multimodal ICL practical \cite{flamin,idefics,qwen,intern}. These advances expand the practical scope of LVLMs \cite{icl1,icl2,re1}.

\begin{figure}[t] 
  \centering                 
  \includegraphics[width=\columnwidth]{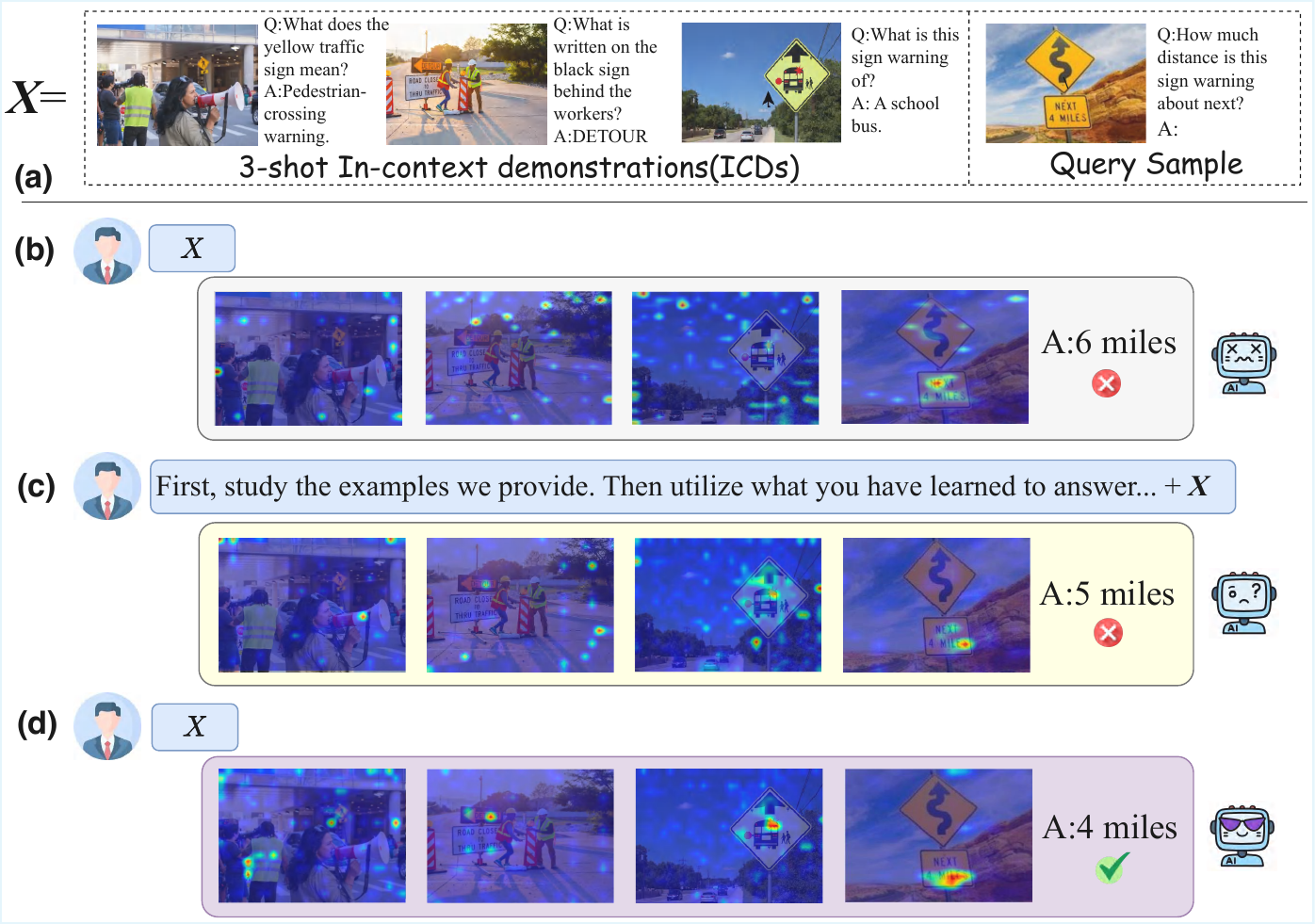}
  \caption{(a) Example of a 3-shot multimodal in-context sequence. (b)-(d) present the vanilla model, adding an instruction to the sequence, and our proposed method, \textbf{CAMA}, respectively. All attention heatmaps come from layer 18, and redder regions indicate stronger attention.}             
  \label{first}           
\end{figure}

However, the benefits of multimodal ICL can be diminished by its pronounced instability. Recent studies have found that LVLM ICL performance is highly sensitive to subtle prompt details. Minor changes in the order or formatting of ICDs can cause significant performance swings \cite{sen1,sen2,sen3}. This sensitivity intensifies as the number of ICDs grows or when complex multimodal reasoning is required \cite{unstable,icv2}. To improve multimodal ICL in LVLMs, two main routes have been explored. The first optimizes the prompt. It adds guiding text, highlights key image regions, or selects and orders ICDs with predefined metrics \cite{conf1, conf2}. However, prompt engineering requires substantial prior knowledge, lacks stability, and depends heavily on the base model. The second route edits models' internal logits, as in contrastive decoding. This approach alters reasoning more directly but needs distorted inputs and extra forward passes for calibration \cite{attn1, contrastive}. Since practical ICL prioritizes efficiency, our work investigates the following questions: \textit{What is the intrinsic limitation underlying the instability of multimodal ICL in LVLMs? Can we address it using a more efficient, training-free method?}

In pursuit of these questions, we examine the attention dynamics of LVLMs during multimodal ICL. Because their inputs interleave images and texts, we move beyond the existing focus on single-image scenarios and analyze LVLMs from two distinct perspectives: (1) text-based visual grounding within each image-text pair, and (2) query-sample-driven attention allocation across ICDs. We visualize attention deficits in both aspects through targeted experiments and verify that these deficits lead to insufficient utilization of the in-context sequence by the LVLM, resulting in unstable multimodal ICL. In response to these deficits, we present Context-Aware Modulated Attention (CAMA), a training-free method that dynamically modulates the model’s internal attention logits during inference based on the input context. CAMA addresses the two deficits with a two-stage modulation that targets the shallow and middle layers of the decoder. Stage I applies \textbf{intra-ICD grounding}, highlighting the image tokens in each ICD and in the query sample that best align with their accompanying text, which reduces the attention sink introduced by the interleaved image-text format. Stage II performs \textbf{query-centric routing}, reallocating attention among ICDs in proportion to their value to produce the desired answer, thus improving the use of context. Experiments demonstrate that CAMA delivers consistent performance gains for multimodal ICL, providing insights for reshaping attention mechanisms to build models with better visual capabilities. 

The contributions of this paper are summarized below:
\begin{itemize}
    \item We analyze LVLMs' attention dynamics in multimodal ICL beyond single-image settings and reveal two deficits: weak vision-text alignment within each pair and misalignment between the query sample and its ICDs.
    \item Building on the identified deficits, we introduce CAMA, the first training-free and model-agnostic method designed to enhance multimodal ICL. CAMA applies a two-stage modulation of internal attention logits that steers the model toward the tokens most relevant to ICL.
    \item Extensive experiments demonstrate that CAMA improves multimodal ICL across diverse LVLMs and benchmarks. Ablation studies confirm the necessity of each design and reveal CAMA’s broader potential.
\end{itemize}

\section{Related Work}

\label{sec:related}

\textbf{In-context Learning (ICL).} ICL enables models to solve unseen tasks by conditioning on an input sequence of input–output examples (i.e., ICDs) without updating any parameters \cite{iclr1,iclr2}. This capability markedly improves their practicality and is especially valued in resource‑intensive multimodal domains \cite{iclr4}. Large vision–language models (LVLMs) gain ICL capabilities through targeted pretraining or fine‑tuning on interleaved image–text data, as in Flamingo \cite{flamin} and LLaVA-NeXT \cite{llava}. Now, it has become a core ability of commercial model families such as QwenVL \cite{qwen} and InternVL \cite{intern}. Consequently, recent work has begun to explore the mechanisms of multimodal ICL \cite{icl1,icl2}. However, few studies probe internal attention. Therefore, cross-modal interactions in multimodal ICL remain only partially understood, which prevents state-of-the-art (SOTA) models from fully exploiting ICL's potential.

\textbf{Enhancing Multimodal ICL.} Limited use of input‑sequence information by LVLMs is considered a major source of instability in multimodal ICL \cite{sen1, sen2, sen4}. Three solution paths have thus emerged. First, task-oriented datasets combined with instruction tuning \cite{tuning1, tuning2} or direct preference optimization (DPO) \cite{DPO} train LVLMs to reason across multiple ICDs. Second, prompt-level methods optimize ICD selection with metrics such as similarity scores and information entropy \cite{conf1,conf2,conf3,order1}, or by training automatic selectors \cite{taco, lever}. Third, calibration-based methods adjust LVLMs' final logits using contrastive decoding \cite{cd1,cd2,cd3}. While these approaches improve performance, they all face challenges. The first requires extensive curated data and parameter updates, the second is less adaptable to fixed-prompt scenarios, and both depend on the base model. The third needs to design distorted input sequences, plus multiple forward passes. In contrast, our CAMA sidesteps all these drawbacks by modulating attention logits at inference time without additional data or tuning.

\section{Attention Dynamics in Multimodal ICL}
\label{mot}
\subsection{Background and Notation}
Current-generation LVLMs generally consist of three core components: a vision encoder that processes images, a projector that converts visual features into embeddings, and an autoregressive LLM that decodes both image and text embeddings to produce output. In multimodal ICL, LVLM typically takes an interleaved image‑text sequence as input, as illustrated in Figure \ref{first}(a). This work focuses mainly on visual question-answering (VQA), which emphasizes both visual perception and language-based reasoning. 

We consider an LVLM $\mathcal{M}$, which generates the answer $y$ given an $n$-shot sequence $X$ as input. $X$ consists of $n$ in-context demonstrations (ICDs) and a single query sample:
\begin{equation}
\begin{aligned}
y \gets&  \mathcal{M}(X),\\
X = (X^{I}_1,\,X^{T}_1,\,\dots,\,&X^{I}_n,\,X^{T}_n,\,X^{I}_{n+1},\,X^{T}_{n+1}),
\end{aligned}
\end{equation}
where the query sample is indexed as $n+1$ for unified notations. $X^{I}_i\in \mathbb{R}^{S_{i}^{I} \times D}$ denotes the token sequence of the $i$-th image, $X^{T}_i\in \mathbb{R}^{S_{i}^{T} \times D}$ denotes the token sequence of the $i$-th text segment and $D$ is the dimensionality of hidden states. $S_{i}^{I}$ and $S_{i}^{T}$ are the token counts of the $i$-th image and text segment, respectively. $S=\sum_{i=1}^{n+1} (S_{i}^{I}+ S_{i}^{T})$ is the total length of the token sequence and $\mathbf{S}$ is its index set. Within each ICD, the text tokens can be further divided into a question part and an answer part, whose counts are $S_{i}^{Q}$ and $S_{i}^{A}$. $\mathbf{S}^{I}_{i}$, $\mathbf{S}^{Q}_{i}$ and $\mathbf{S}^{A}_{i}$ denote the token index sets for the image, question, and answer of the $i$-th ICD, respectively.

The sequence $X$ is then passed to the LLM of $\mathcal{M}$, which performs an $N$-layer decoder forward pass. Each layer contains a multi-head attention (MHA) module. The $h$-th head in the $l$-th layer maps the hidden states of $X$ to queries $Q^{l,h} \in \mathbb{R}^{S\times D_{k}}$, keys $K^{l,h} \in \mathbb{R}^{S\times D_{k}}$, and values $V^{l,h} \in \mathbb{R}^{S\times D_{k}}$ by linear transformations, where $D_{k}$ is the head dimension. Attention logits $\mathbf{A}^{l,h} \in \mathbb{R}^{S\times S}$ is given by:
\begin{equation}
\label{attneq}
\mathbf{A}^{l,h}
  =
      \frac{Q^{l,h}\,(K^{l,h})^{\top}}
           {\sqrt{D_k}}
    ,
\end{equation}
which directly reveals both the direction and intensity of interactions between any two tokens during the forward pass. After applying a causal mask followed by softmax, $\mathbf{A}^{l,h}$ becomes the attention weight matrix that is multiplied with the value $V^{l,h}$ to produce the head's attention output.

\subsection{Attention Deficits of LVLMs}
\begin{figure}[t] 
  \centering                 
  \includegraphics[width=0.9\columnwidth]{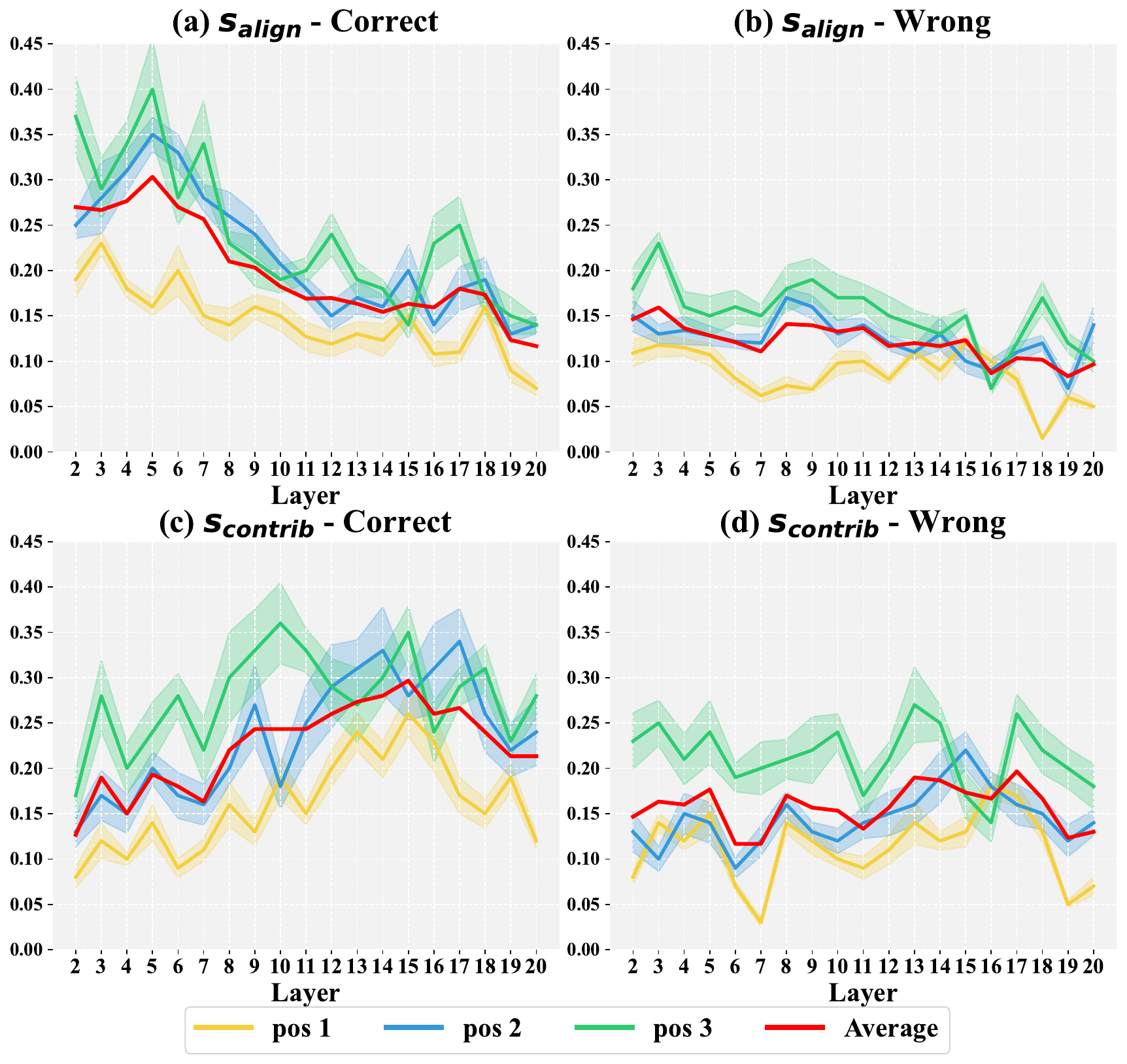}
  \caption{Layer-wise trends of the intra-ICD alignment score $s_{align}$ and and the key ICD contribution score $s_{contrib}$ in effective and ineffective multimodal ICL. Pos 1, 2, and 3 denote the key ICD position in the sequence.}             
  \label{deficit}           
\end{figure}
To understand the instability of multimodal in-context learning in LVLMs, we delve into the model's attention mechanism to analyze its dynamics in both effective and ineffective scenarios. Our objective is to uncover distinct patterns that distinguish these cases. Specifically, we investigate attention dynamics at two levels: (1) Within each ICD: We examine text-based visual grounding to verify whether the model pays attention to relevant objects in the images. (2) Across ICDs: We evaluate whether the model effectively distributes attention among different ICDs according to their relevance to the query. To ensure generalizable experimental results, we experiment with \textbf{3}-shot settings on two LVLMs: Llava-NeXT-7B \cite{llava} and Idefics2-8B \cite{idefics2}. Query samples and ICDs are taken from the validation and training sets of VQAv2, respectively.

\textbf{Setups.} 
We pair each query with three ICDs of the same question type (e.g., ``How many” or ``Is there") and process the resulting sequences using both LVLMs. From this pool, we identify 2,500 sequences where LLaVA-NeXT-7B produces a correct answer while Idefics2-8B fails, and another 2,500 sequences with the opposite outcome. These form two distinct sets of 5,000 cases each: a ``correct'' group (indicating effective ICL) and a ``wrong'' group (indicating ineffective ICL). For each sequence, we manually annotate bounding boxes on the ICD images based on their corresponding textual descriptions. We then extract attention maps from each model layer, identify the top 20\% most-attended regions, and compute their Intersection over Union (IoU) with the annotated boxes to obtain the \textbf{intra-ICD alignment score} \( s_{\text{align}} \). Additionally, we sample 10,000 sequences from the original pool and replace two of the three ICDs in each sequence with unrelated images, leaving only one key ICD relevant to the query type. For each of these sequences, we generate three variants by positioning the key ICD in the first, second, and third slots, respectively. Using the same processing pipeline, we construct new correct and wrong groups of 5,000 cases each. For these, we compute saliency maps~\cite{sail} at each layer and measure the proportion of information flow from the key ICD to the generated answer tokens, relative to the total layer-wise flow. This yields the \textbf{contribution score} \( s_{\text{contrib}} \). Experimental details are provided in Appendix 2.

\textbf{Results.} As shown in Figure~\ref{deficit}, in the correct group, LVLMs exhibit significantly higher \( s_{\text{align}} \) in the shallow layers (Layers 2--4), with the largest score gap compared to the wrong group also occurring in these early layers. This suggests that effective ICL relies on aligning visual attention with textual semantics from the outset.  Furthermore, in the correct group, \( s_{\text{contrib}} \) shows a marked increase beginning around Layer 10 and remains consistently high through Layer 20. In contrast, this rise is absent in the wrong group, where scores remain consistently lower. This indicates that mid-layer attention allocation—driven by the relevance of the query—is also critical for successful ICL. We refer to the failure of LVLMs to establish either of these attention dynamics as \textbf{attention deficits}, which contribute to instability in in-context learning. Additionally, we observe that the earlier an ICD appears in the sequence, the lower its alignment and contribution scores across all layers and both groups.

We distill three core findings from the above results:
\begin{itemize}
    \item \textbf{\textit{Finding 1}}: In shallow layers, LVLMs struggle to focus on visual information that aligns with the text semantics within each image-text pair.
    \item \textbf{\textit{Finding 2}}: In middle layers, LVLMs struggle to prioritize the key ICDs that match the query.
    \item \textbf{\textit{Finding 3}}: Both deficits are amplified by the position of ICDs, with those earlier in the sequence experiencing more severe impacts.
\end{itemize}

\section{Method} 
\label{sec:method}

\subsection{Overview}
\begin{figure}[t] 
  \centering                 
  \includegraphics[width=1\columnwidth]{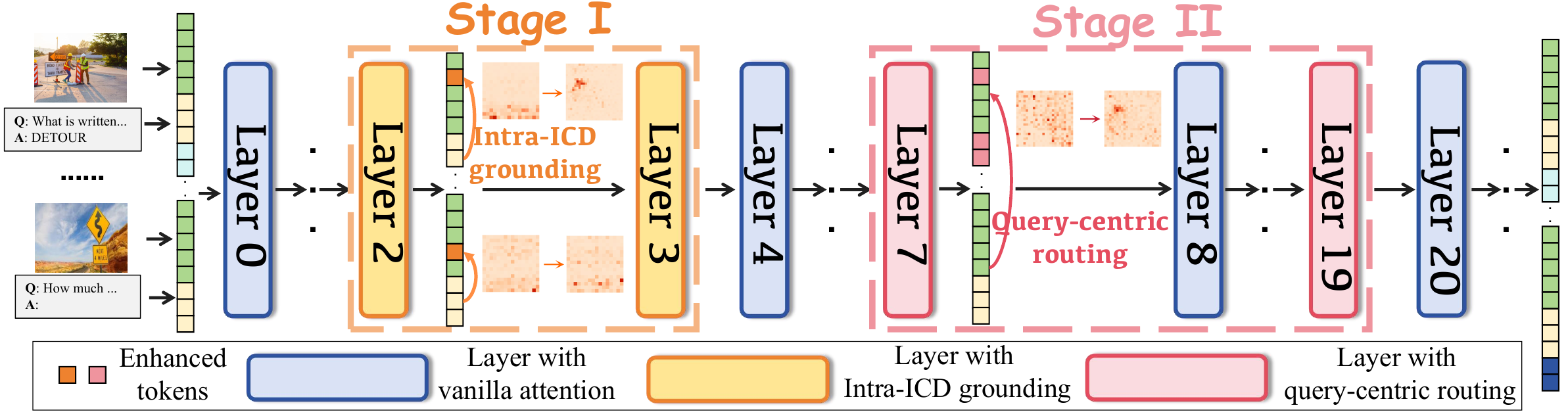}
  \caption{An overview pipeline of CAMA. A version with more details is provided in Appendix 1.}             
  \label{main}           
\end{figure}

To mitigate the two attention deficits at inference time without training, we introduce \textbf{Context‑Aware Modulated Attention} (CAMA). CAMA dynamically reshapes the internal attention logits according to the input sequence during the prefilling process, encouraging the model to focus on the tokens most key to effective ICL. The overall pipeline is depicted in Figure \ref{main}. Building on \textbf{\textit{Finding 1}} and \textbf{\textit{2}}, we divide CAMA into two stages, each attached to a different depth of the LLM and mainly aimed at a specific deficit: 
\begin{itemize}
    \item \textbf{Stage I:} In the shallow layers, CAMA performs \textbf{Intra-ICD grounding}. At this stage, for each ICD, we first locate the key image tokens that are essential to justify the provided answer. We then amplify the attention paid to these tokens so the model captures the critical visual cues, laying a solid foundation for subsequent ICL.
    \item  \textbf{Stage II:} In the middle layers, CAMA performs \textbf{query-centric routing}. At this stage, we operate at the attention head level to manage the complex interactions between the query sample and the ICDs. Specifically, we identify the heads that exhibit the strongest query-to-ICD attention and rescale their logits based on the cross-modal similarity between the query and each ICD.
\end{itemize} 

\subsection{Stage I: Intra‑ICD grounding}
Stage I operates on the shallow layer set \(\mathcal{L}_{\text{stageI}}\), focusing on improving image-text alignment within each ICD to enhance LVLM’s early perception of key visual features. This not only mitigates the early attention deficits but also benefits subsequent layers. To achieve this goal, we first need to identify the key visual tokens in each ICD based on the semantics of its paired Q-A text. However, conventional metrics, such as summing attention scores from text to image tokens or relying on embedding similarity, are inadequate for multimodal ICL, as they may exacerbate attention deficits or fail to capture the nuanced semantics of Q-A dialogues~\cite{qa}. To address this, we introduce a \textbf{dynamic attention increment} strategy tailored to multimodal ICDs.

For the $i$-th ICD in $X$, we designate three anchor tokens $\mathbf{T}_i$: the first token of the question $\mathbf{S}^{Q}_{i}[0]$, the first token of the answer $\mathbf{S}^{A}_{i}[0]$ and the last token of the answer $\mathbf{S}^{A}_{i}[-1]$. In our setting, these tokens are typically ``Q'', ``A'', and a punctuation mark (as illustrated in Figure \ref{first}(a)). After the first layer, the forward pass enables these tokens to capture and summarize the semantics of the preceding tokens. Therefore, they can be treated as proxies for the overall semantics of the image, the question, and the answer. We begin by computing the attention distributions that these anchor tokens assign to every image token $j\in\mathbf{S}^{I}_{i}$ in layer $l\in\mathcal{L}_{\text{stageI}}$:
\begin{equation}
\small
P_l(\mathbf{T}_i)=\operatorname{softmax}_{j}\Bigl(\tfrac{1}{H}\sum_{h=1}^{H}\textbf{A}^{l,h}(\mathbf{T}_i,j)\Bigr)\in\mathbb{R}^{1\times S^{I}_{i}} ,
\end{equation}
where $\mathbf{T}_i\in \{\mathbf{S}^{Q}_{i}[0], \mathbf{S}^{A}_{i}[0], \mathbf{S}^{A}_{i}[-1]\}$ denotes the anchor tokens. The inner summation averages the logits over all $H$ heads in $l$, and the row‑wise softmax turns this average into a probability distribution across the image tokens.

Next, we calculate the differences between these distributions as dynamic attention increments. They provide bias-reduced estimates of the image tokens' contributions to both question understanding and answer making within each ICD. We quantify them by defining two non-negative forward gains $c^{1}_{l,i},c^{2}_{l,i}\in\mathbb{R}^{1\times S^{I}_{i}}$in a divergence-like form:
\begin{align}
\small
c^{1}_{l,i}&=
\bigl[P_l(\mathbf{S}^{A}_{i}[0])\!-P_l(\mathbf{S}^{Q}_{i}[0])\bigr]_{+}\,
\log\frac{P_l(\mathbf{S}^{A}_{i}[0])}{P_l(\mathbf{S}^{Q}_{i}[0])},
\\
c^{2}_{l,i}&=
\bigl[P_l(\mathbf{S}^{A}_{i}[-1])\!-P_l(\mathbf{S}^{A}_{i}[0])\bigr]_{+}\,
\log\frac{P_l(\mathbf{S}^{A}_{i}[-1])}{P_l(\mathbf{S}^{A}_{i}[0])},
\end{align}
where $[\cdot]_{+}$ keeps only the positive values by setting all negative ones to zero. After summing the two gains, column $j$ is the score of the $j$-th image token:
\begin{equation}
s_{i,j}=\sum_{l\in\mathcal{L}_{\text{stageI}}}^{} (c^{1}_{l,i}+c^{2}_{l,i})[j].
\end{equation}

A larger $s_{i,j}$ means the token plays a greater role in the visual perception of the image by LVLM based on the corresponding Q-A pair and therefore is more closely aligned with the text semantics. For each $i$-th ICD, we take the image token indices with the top-$k_\text{I}\%$ scores and define them as the key set $\mathcal{K_{I}}_{i}$. Finally, we modulate the attention logits in Eq.\ref{attneq} to amplify the key image tokens' \emph{incoming} attention:
\begin{equation}
\textbf{A}^{l,h}(r,j)\leftarrow\
\textbf{A}^{l,h}(r,j)+
\,
\frac{n-i+1}{n}\,
\frac{s_{i,j}}{\displaystyle\max_{j'\in\mathcal{K_{I}}_{i}}s_{i,j'}+\epsilon}.
\label{stage1}
\end{equation}
where $j\in \mathcal{K_{I}}_{i}$ and $r$ denotes any later token in $X$. The factor $(n-i+1)/n$ is used to offset the position bias noted in \textbf{\textit{Finding 3}}. Note that during this stage, we also detect and enhance the query sample’s key image tokens through only $c^{1}_{l,n+1}$. With intra‑ICD grounding, image tokens that align well with corresponding textual semantics receive greater attention in subsequent layers and during answer generation, providing a bias-corrected basis for deeper reasoning.

\subsection{Stage II: Query‑centric routing}
After highlighting the most semantically aligned image tokens in Stage I, Stage II works on the middle layers $\mathcal{L}_{\text{stageII}}$ to refine the global information flow under the guidance of the query. Inspired by \cite{head}, which shows that specific attention heads dominate the information flow from the query sample to ICDs during ICL, we perform fine‑grained and targeted modulation at the head level to lessen the influence of other complex interactions in multimodal ICL. 

We first use the attention flow from the query sample to the context (i.e., ICDs) as a signal to identify \textbf{query-centric heads}. As information gradually shifts toward later tokens in the middle layers, and text typically exhibits a stronger attention‑flow tendency \cite{flow}, we measure the \emph{query→context flow} using only the query sample's text. Recall that $\mathbf{S}^{T}_{n+1}$ denotes the text token index set of the query sample. For each head $h$ in layer $l\!\in\!\mathcal{L}_{\text{stageII}}$, we compute:
\begin{equation}
\rho^{l,h}=
\frac{1}{|\mathbf{S}^{T}_{n+1}|}
\sum_{q\in\mathbf{S}^{T}_{n+1}}
\sum_{c\in\mathbf{S}_{ctx}}
\textbf{A}^{l,h}(q,c),
\label{eq:rho}
\end{equation}
where $\mathbf{S}_{\text{ctx}}\!=\!\bigcup_{i=1}^{n}\!\bigl(\mathbf{S}^{I}_{i}\!\cup\!\mathbf{S}^{Q}_{i}\!\cup\!\mathbf{S}^{A}_{i}\bigr)$. $\rho_{l,h}$ aggregates the attention flow from the query sample to the preceding context, allowing us to quantify each head’s contribution in that direction. The heads are then ranked by $\rho_{l,h}$, and the top-$k_\text{II}\%$ are chosen as query-centric heads, forming the set $\mathcal{H}_{l}^{\text{QC}}$.

Next, we perform ICD‑level attention modulation within each query-centric head. In the middle layers, token‑level attention becomes blurred by aggregation and is suboptimal for separating the contributions of individual ICDs. Thus, we propose a similarity-based method. To balance semantic maturity and clarity, the hidden states from the final layer of Stage I, $\mathcal{L}_{\text{stageI}}[-1]$, are used for subsequent computations. 

For the embeddings of the $i$‑th ICD taken from $\mathcal{L}_{\text{stageI}}[-1]$, we compute the mean of the key image tokens $\mathcal{K_{I}}_{i}$ to obtain a visual vector and the mean of all question‑and‑answer tokens to obtain a textual vector. We then concatenate these two vectors and apply $\ell_2$-normalization to the resulting embedding, yielding the joint representation $p_i$. We apply the same steps to the query sample to obtain $p_{\text{query}}$. The cosine similarity between $p_i$ and $p_{\text{query}}$ is the query-centric score:
\begin{equation}
w_{i}=
\frac{\exp\!\bigl(\langle p_{i},p_{\text{query}}\rangle\bigr)}
     {\sum_{k=1}^{n}\exp\!\bigl(\langle p_{k},p_{\text{query}}\rangle\bigr)}.
\end{equation}

The score $w_i$ quantifies how semantically relevant each ICD is to the query sample. Applying this score in the query‑centric heads mitigates the LVLM’s difficulty in locating crucial contexts and improves answer generation. Specifically, for each head $h\in\mathcal{H}_{l}^{\text{QC}}$ in layer $l \in \mathcal{L}_{\text{stageII}}$, we modulate the attention logits in Eq.\ref{attneq} as follows:
\begin{equation}
\textbf{A}^{l,h}(r,J)\;\leftarrow\;
\textbf{A}^{l,h}(r,J)+
\frac{n-i+1}{n}\,
w_{i}.
\end{equation}

We apply this modulation to all key image tokens and text tokens of each ICD to maintain completeness, so $J \in (\mathcal{K_{I}}_{i} \cup \mathbf{S}^{Q}_{i} \cup \mathbf{S}^{A}_{i})$ and $r$ denote any subsequent token in $X$. $(n - i + 1)/n$ applies the same position decay as in Eq.\ref{stage1}. Stage II enhances the attention given to each ICD in proportion to its contribution, ensuring that key information is not lost within the extensive context. The two stages of CAMA jointly enable the LVLM to exploit the provided context more effectively. All remaining layers maintain their original attention.

\section{Experiments}
\input{maintab}
\subsection{Setup}

\noindent\textbf{Benchmarks and models.} Following the standard multimodal ICL evaluation \cite{openflamingo}, we test CAMA on VQAv2 \cite{vqav2}, VizWiz \cite{VizWiz}, and OK-VQA \cite{OK-VQA}. To further assess the generalization of CAMA, we also evaluate it on GQA \cite{GQA}, TextVQA \cite{textvqa}, the CLEVR subset of VL-ICL bench \cite{vlicl}, and MMStar \cite{mmstar}. In addition to LLaVA-NeXT-7B and Idefics2-8B, we also report results on two latest LVLMs, InternVL2.5-8B \cite{intern} and Qwen2.5VL-7B \cite{qwen}.

\noindent\textbf{Baselines.} We compare CAMA with five baselines. (1) \textbf{Vanilla} denotes the vanilla models. (2) The instruction‑augmented method (\textbf{+Inst}) add an instruction before each sequence: ``First, study the examples we provide. Then utilize what you have learned to answer the new question.'' (3) Contrastive decoding (\textbf{CD}) \cite{contrastive} replaces each ICD image with a blank one and uses the distorted logits to calibrate the original logits. (4) Visual enhancement (\textbf{VE}) \cite{enhance} manually draws a red bounding box around the relevant region of each ICD image. (5) SoFt Attention (\textbf{SoFA}) \cite{sofa} is a training‑free method that inserts a bidirectional attention mask after every two decoder layers, which reduces position bias when multiple images are contained in the input.

Appendix 3.1 provides further introductions to the benchmarks, models, baselines, and processing details.

\noindent\textbf{Implementation details.} For each benchmark, samples in its validation set act as query samples, each paired with \textbf{eight} randomly retrieved ICDs from the training split, forming an 8-shot sequence. Stage I is applied to the 2nd and 3rd layers. Stage II is applied to every second layer from the 7th through the 19th. We set $k_\text{I}=k_\text{II}=20$. All experiments are conducted on NVIDIA H200 GPUs.
\subsection{Main Results}
\noindent\textbf{CAMA is effective and robust across all VQA benchmarks and LVLMs.} Table \ref{tab:main} presents accuracy results across seven VQA benchmarks for four LVLMs with varying input image resolutions and LLM decoders. CAMA achieves the highest accuracy in all 28 experiments, surpassing all baselines. On average it raises accuracy over the vanilla models by 2.96\%. Notably, stronger models benefit even more: InternVL2.5 and Qwen2.5VL see improvements of 3.61\% and 3.15\%, respectively, compared to 2.35\% on LLaVA-NeXT and 2.73\% on Idefics2. These findings demonstrate the strong effectiveness and generalization of CAMA. Thanks to its plug-and-play design, CAMA can consistently benefit the emerging open-source LVLMs, giving it promising practical value. Moreover, by outperforming SoFA’s mask-based strategy, CAMA confirms the advantage of modulating intermediate attention logits.

\noindent\textbf{CAMA can activate the effect of prompt-based methods.} We also report results that combine CAMA with a prompt-based baseline, as shown in Table \ref{tab:main} in the rows ``CAMA(+Inst)'' and ``CAMA(VE)''. We find that a prompt-based method alone gives only a marginal performance gain, which confirms the attention deficits inside LVLMs. When we add CAMA the model not only improves on its own but also \textbf{activates} the real benefit of these methods. For example, +Inst exceeds the vanilla model by just 0.27\%, whereas CAMA lifts this improvement to 3.28\% and adds another 0.32\% over using CAMA alone. The performance gain brought by CAMA's activation is most evident on CLEVR, where VE sharply reduces the difficulty of cognizing the original image-text mapping. This result further confirms CAMA's practical promise, as it allows curated prompts to exert their full effect and maximizes model performance.

\begin{table}[t]
  \centering
  \begingroup
    \setlength{\tabcolsep}{1.6pt}
    \small
  \begin{tabular}{ccccc}
    \toprule
    \multirow{2}{*}{\textbf{Method}} & \multicolumn{2}{c}{\textbf{Image captioning}} & \textbf{Classification} & \textbf{Storytelling}\\
    & Flickr30k & MSCOCO  & Hatefulmemes & L-I-VST\\ 
    \midrule
    Vanilla & 69.93 & 112.46 &  75.16 & 38.61\\
    +Inst   &  71.28  &   114.36  &   76.20   & 38.46\\ 
    CD      &   72.35    &  114.97   &   74.38    & 40.32\\ 
    SoFA     & 72.65 & 115.89 & 76.51& 40.41\\
    \rowcolor{lightgray}
    CAMA    & \underline{73.88} & \underline{116.72}&\underline{77.49} & \underline{41.79}\\ 
    \rowcolor{lightgray}
    CAMA(+Inst) & \textbf{74.37}& \textbf{117.04}& \textbf{77.93}& \textbf{42.38}\\ 
    \bottomrule
  \end{tabular}
  \caption{Average performance of 8-shot multimodal ICL on four benchmarks spanning three additional tasks, reported using CIDEr↑ , ROC-AUC↑, and L-I-score↑, respectively.}
  \label{tab:gen}
  \endgroup
\end{table}

\noindent\textbf{CAMA exhibits strong cross-task generalization.} To fully evaluate the generalization of CAMA, we introduce three additional tasks beyond VQA: image captioning, image classification, and visual storytelling. In these tasks, the text paired with each image is no longer a Q-A pair; instead, it is a caption, a class label, or a narrative sentence, and all three start with the prefix ``A:''. The text part of the query sample only contains ``A:''. Therefore, for these tasks, Stage I of CAMA is based solely on attention gains $c^{1}_{l,i}$ to identify key image tokens. The results are reported in Table \ref{tab:gen}. CAMA again achieves superior performance on all these tasks and activates the gains of +Inst, showing that its benefits extend beyond VQA to broader interleaved scenarios.

\subsection{Ablation Study and Analyses}

\begin{figure}[t] 
  \centering                 
  \includegraphics[width=0.85\columnwidth]{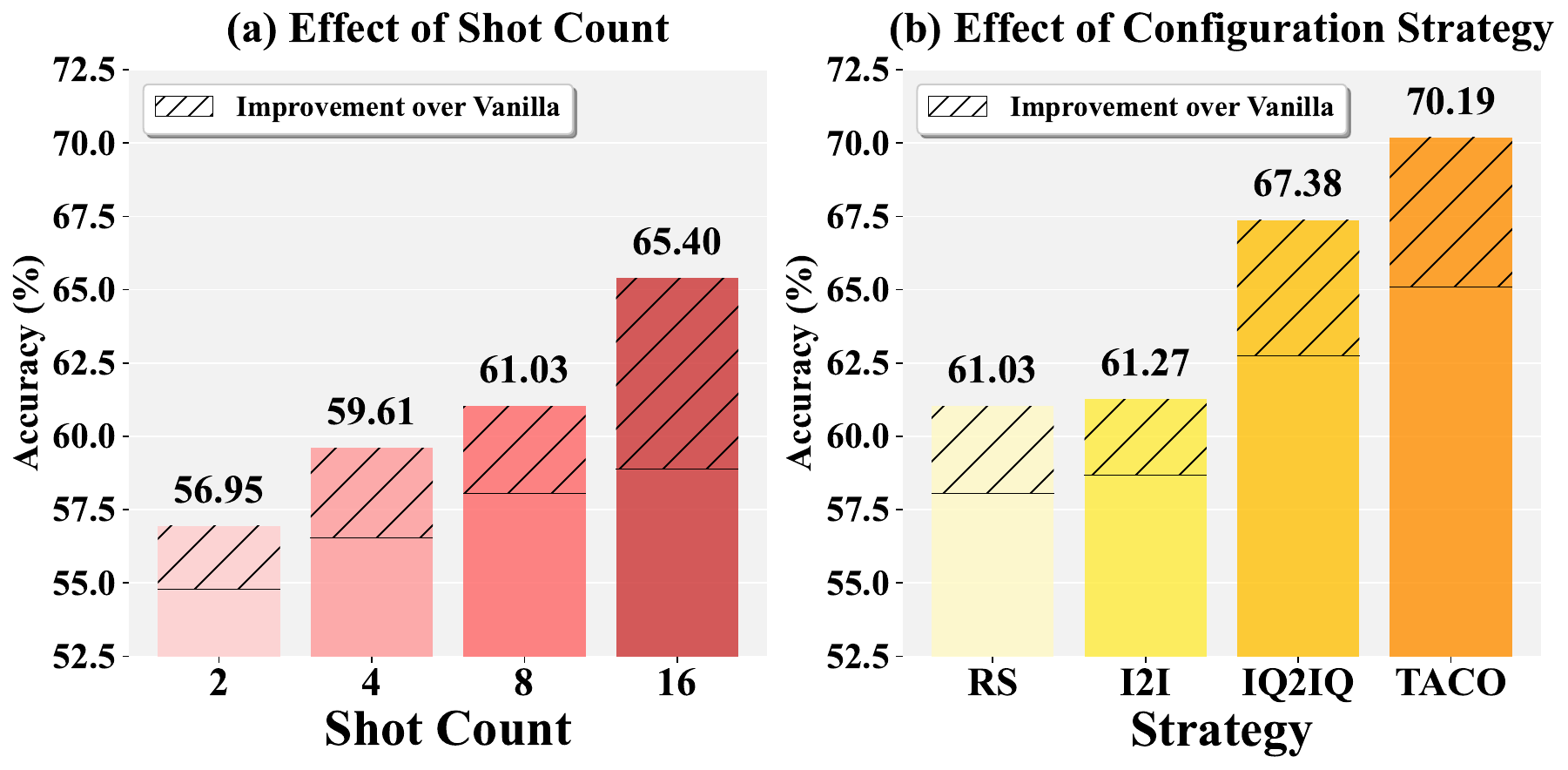}
  \caption{Average performance of CAMA across four LVLMs and seven VQA benchmarks as the count of ICDs and sequence configuration strategies vary.} \label{abla1}          
\end{figure}

\textbf{Adaptability to diverse ICD counts.} It is non-trivial for enhancement methods to adapt to different shot counts to meet various practical needs. As Figure \ref{abla1}(a) shows, CAMA consistently outperforms the vanilla model in 2, 4, 8, and 16-shot configurations, yielding gains of 2.15\% to 6.52\%. These results validate CAMA’s generalization across diverse scenarios and its ability to meet varying requirements. Meanwhile, CAMA’s advantage grows as the sequence length increases, which aligns with \textbf{\textit{Finding 3}} because longer contexts intensify positional bias. As LVLM context windows expand and many‑shot ICL becomes feasible, CAMA provides a promising path for further progress.

\noindent\textbf{Adaptability to diverse sequence configurations.} The in‑context sequence configuration is key to ICL performance. Our main experiments build sequences using uniform random sampling (RS), whereas some advanced systems leverage embedding similarity to improve sequence quality further. Each candidate ICD and the query sample are encoded using CLIP-ViT-L/14 to compute cosine similarity, and the top eight matches are selected as ICDs. I2I uses image embeddings only, while IQ2IQ concatenates image and question embeddings. We also adopt TACO \cite{taco}, a SOTA language-model-based ICD retriever. As shown in Figure \ref{abla1}(b), CAMA consistently improves accuracy by 2.59\% to 5.08\%. Stronger retrieval brings larger gains, indicating that CAMA’s activation effect also applies to configuration methods. This compatibility allows CAMA to integrate smoothly with future ICL advances.

\begin{table}[t]
\centering
\begingroup
\setlength{\tabcolsep}{1.6pt}
\footnotesize
\begin{tabular}{cccccccc}
\toprule
\textbf{Method} & \textbf{VQAv2} & \textbf{VizWiz} & \textbf{OK-VQA} & \textbf{GQA} & \textbf{TextVQA}\\ 
\midrule
CAMA & 68.12 & 50.69 & 60.56 & 68.60 & 77.53 \\
\midrule
w/o Stage I & 67.05 & 49.23 & 59.46 & 68.09 & 76.99 \\
w/o Stage II & 66.41 & 48.87 & 58.40 & 67.21 & 76.08 \\
\midrule
\multicolumn{6}{c}{Stage I}\\
w/o increment & 67.14 & 48.86 & 59.28 & 67.39 & 76.80 \\
w/o top-$k_\text{I}\%$ & 67.44 & 50.10 & 59.82 & 67.93 &77.14\\
w/o position & 67.91 & 50.12 & 60.02 & 68.05 & 76.98 \\
\midrule
\multicolumn{6}{c}{Stage II}\\
w/o top-$k_\text{II}\%$ & 67.18 & 49.25 & 59.84 & 68.11 & 76.69 \\
$L_\text{stageI}[0]$ & 67.85 & 49.16 & 60.28 & 68.20 & 77.05 \\
w/o position & 68.03 & 49.84 & 60.19 & 68.20& 77.12 \\
\bottomrule
\end{tabular}
\caption{Average 8-shot performance of CAMA on four LVLMs under different ablation settings. ``w/o increment'' disables dynamic attention increment and computes the attention score using only $\mathbf{S}^{A}_{i}[-1]$. ``w/o top $k_\text{I}\%/k_\text{II}\%$'' applies CAMA to all image tokens or heads. ``$L_\text{stageI}[0]$'' computes similarity with embeddings from $L_\text{stageI}[0]$. ``w/o position'' removes the position-decay factor.}
\label{tab:abla}
\endgroup
\end{table}

\noindent\textbf{The superiority of CAMA designs.} We conduct a comprehensive ablation study on the two stages of CAMA and the key components within each stage (Table \ref{tab:abla}). Removing either Stage I or Stage II leads to a noticeable drop in performance, with the absence of Stage II resulting in a more substantial degradation. This confirms that the two-stage modulation is necessary and that stronger query-guided reasoning is more crucial for multimodal ICL. Disabling each key design of a stage also produces different levels of degradation. We draw the following conclusions: (1) Selecting the key image tokens is essential when enhancing image tokens in shallow layers. It prevents our modulation from being diluted by the later softmax. (2) Query-centric heads play a vital role in multimodal ICL. Targeted enhancement of these heads promotes specific information flows. (3) The dynamic attention increment cleverly avoids the bias introduced by the attention sink. (4) The position-decay factor in CAMA effectively alleviates position bias when LVLMs perform ICL with long sequences, preventing crucial information at the beginning of the sequence from being overlooked. This is particularly important in multimodal ICL, where each image consumes a large number of tokens and leads to long contexts, thereby improving overall performance. Intuitively, as the number of ICD shots increases, the role of the position factor becomes more significant, as discussed in Appendix 3.2.

\begin{figure}[t] 
  \centering                 
  \includegraphics[width=0.85\columnwidth]{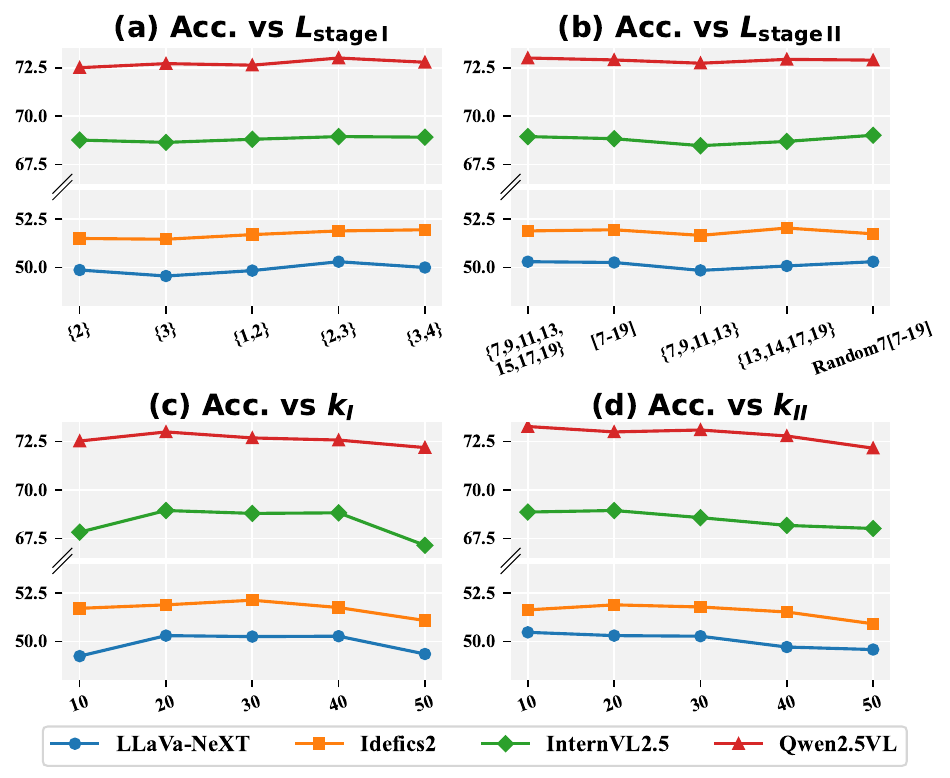}
  \caption{Average performance of CAMA on seven VQA benchmarks while varying $L_\text{stageI}$, $L_\text{stageII}$, $k_\text{I}$, and $k_\text{II}$. Random7 means randomly choosing seven layers.} 
  \label{abla2}          
\end{figure}

\noindent\textbf{Impact of the selection of layers and hyperparameters.} As shown in Figure \ref{abla2}, the performance of CAMA remains stable as we vary the layers where the two stages are applied. Meanwhile, when $k_\text{I}$ is set between 20\% and 40\%, CAMA stays consistent, and when $k_\text{II}$ is between 10\% and 30\%, CAMA also stays consistent. This behavior indicates that query-centric heads are more targeted than key image tokens, yet both possess a relatively wide optimal range. The results demonstrate that CAMA is robust to layer and hyperparameter choices and does not require carefully curated tuning strategies, highlighting its practicality and efficiency.

\noindent\textbf{Case Study.} In the case of Figure \ref{first}, we observe that CAMA reshapes attention, guiding the model to focus on visual features aligned with the text and relevant to the query. This shows the depth of CAMA’s ability to address attention deficits. We provide additional qualitative visualizations and failure case analysis in Appendix 3.4. For challenging tasks like CLEVR with vague text, Stage I's dynamic attention increment helps locate key tokens. For MMStar, which involves complex long-text tasks, CAMA enables deeper reasoning. This highlights CAMA’s potential in complicated or reasoning-intensive scenarios, such as multimodal Chain-of-Thought (CoT) \cite{cot}.

\section{Conclusion}
We introduce CAMA, a training-free and plug-and-play attention modulation method that enhances multimodal ICL. We begin by analyzing the attention dynamics of LVLMs in this setting and, guided by the findings, design a two-stage modulation pipeline for CAMA. CAMA effectively corrects attention deficits and guides the model to focus on image regions that contribute the most to multimodal ICL. Experiments across multiple benchmarks and LVLMs show that CAMA achieves superior results. We believe that CAMA can inspire new directions for future advances of LVLM.

\noindent\textbf{Limitations.} As CAMA needs to modulate LVLM's attention, it introduces additional latency compared with the vanilla model. However, this latency is acceptable, especially in light of the performance gains brought by CAMA. An efficiency analysis is provided in Appendix 3.5.

\section*{Acknowledgments}
We acknowledge the computing resources provided by NSF ACCESS.

\bibliography{aaai2026}
\section{Appendix}
\subsection{1 $\quad$ Figure of Pipeline}
\begin{figure*}[t] 
  \centering                 
  \includegraphics[width=\textwidth]{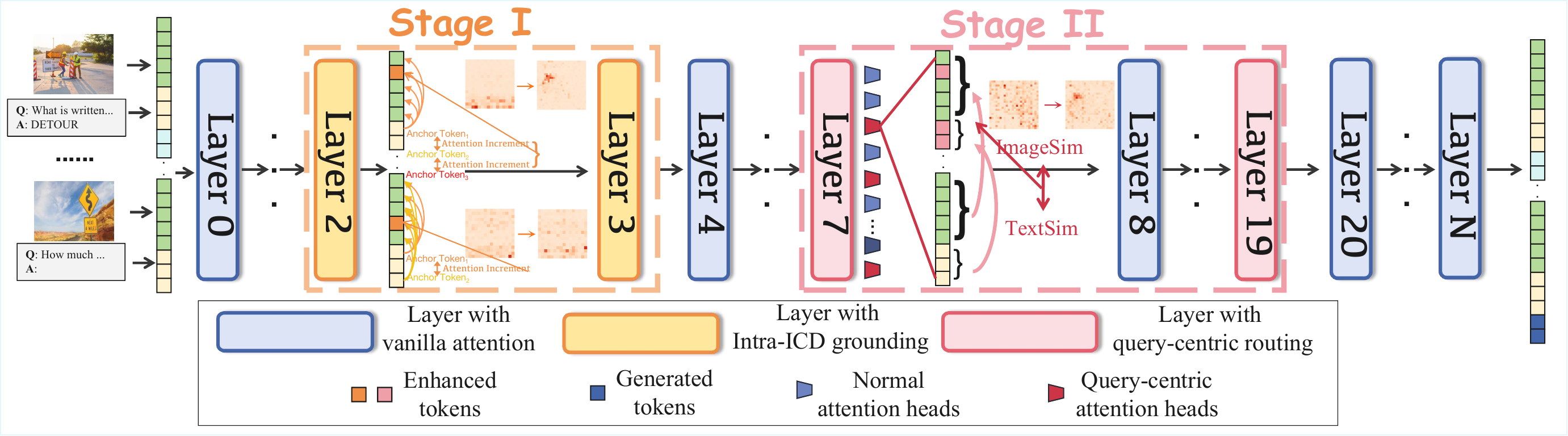}
  \caption{Overview pipeline of CAMA. CAMA consists of two stages. Stage I is applied to the shallow layers of the LVLM decoder. At this stage, CAMA performs intra-ICD grounding. For each ICD, CAMA first locates the key image tokens that are essential to justify the provided answer by computing dynamic attention increments from the anchor tokens to every image token, and then amplifies the attention paid to these tokens so the model captures the critical visual cues. Stage II is applied to the middle layers of the LVLM decoder. In this stage, CAMA operates at the level of attention heads to manage the complex interactions between the query sample and the ICDs. It identifies the heads that exhibit the strongest query-to-ICD attention and rescales their logits based on the cross-modal similarity between the query sample and each ICD.} 
  \label{main_d}           
\end{figure*}

We present a detailed version of CAMA's pipeline in Figure \ref{main_d}.

\subsection{2 $\quad$ Experiments on Attention Deficits in LVLMs}
\paragraph{Query–ICD Pairing and Dataset Construction.}
We conduct experiments on the VQAv2 dataset. In the training set, each instance consists of an image, a question, and an answer. The validation set does not include answers. We treat every example in the validation set as a query sample. For each query sample, we randomly retrieve three instances with the same question type from the training set based on the annotated question type (e.g.\ \textit{"Is there"} and \textit{"How many"}). These instances serve as ICDs for the given query sample. The resulting 4-tuple  
\begin{equation}
s_i=\bigl\{\text{ICD}_{i,1},\;\text{ICD}_{i,2},\;\text{ICD}_{i,3},q_i\bigr\}
\end{equation}
forms a homogeneous 3-shot in-context sequence.  
This yields a corpus $\mathcal{S}=\{s_i\}_{i=1}^{N}$ whose size $N$ matches VQAv2's validation set.

\paragraph{Model Inference and Correct / Wrong Split.}
Each sequence $s_i$ is fed into two LVLMs, LLaVA-NeXT-7B ($\mathcal{M}_1$) and Idefics2-8B ($\mathcal{M}_2$), producing answers $\hat a_i^{(1)}$ and $\hat a_i^{(2)}$, respectively. Comparing them with the ground-truth $a_i^\star$, we collect
\begin{equation}
\mathcal{C}_{1}=\bigl\{s_i\!\mid\!\hat a_i^{(1)}=a_i^\star,\;\hat a_i^{(2)}\neq a_i^\star\bigr\},
\end{equation}
\begin{equation}
\mathcal{C}_{2}=\bigl\{s_i\!\mid\!\hat a_i^{(1)}\neq a_i^\star,\;\hat a_i^{(2)}=a_i^\star\bigr\}.
\end{equation}

We randomly sample $2{,}500$ sequences from each of $\mathcal{C}_{1}$ and $\mathcal{C}_{2}$, and each sequence corresponds to one correct case and one wrong case. Consequently, we obtain two case groups: the \emph{correct group}  
$\mathcal{G}_{\text{correct}}$  
and, symmetrically, a \emph{wrong group} $\mathcal{G}_{\text{wrong}}$ of the same size (5,000 each). We use $\mathcal{G}_{\text{correct}}$ to represent effective ICL and $\mathcal{G}_{\text{wrong}}$ to represent ineffective ICL. 

\paragraph{Intra-ICD Alignment Score $s_{\text{align}}$.}
For every ICD position $p\!\in\!\{1,2,3\}$ within a sequence $s_i$, we manually annotate the regions in its image that are related to the semantics of the corresponding Q–A pair with red bounding boxes, denoted as $\mathcal{B}_{i,p}$. Each $\mathcal{B}_{i,p}$ contains the visual cues necessary to derive the answer from the question. We conduct two additional human verification passes on the annotations to ensure their reliability. At decoder layer $l$ we extract the attention matrix $\mathbf{A}_{i,l}\!\in\!\mathbb{R}^{S^{\text{total}_i}\times S^{\text{total}_i}}$, where $S^{\text{total}_i}$ denotes the total length of $s_i$ and the generated tokens. We convert it to a heat map $\mathbf{H}_{i,l}$ via row-wise $\max$–normalization. Let $\mathcal{V}_{i,p}$ denote the set of visual tokens of $\text{ICD}_{i,p}$. We extract the top-20\% highest-valued tokens within $\mathcal{V}_{i,p}$ to form the salient region:
\begin{equation}
R_{i,p,l}
\;=\;
\operatorname*{Top-20\%}_{v\in\mathcal{V}_{i,p}}\!\bigl(\mathbf{H}_{i,l}(v)\bigr).
\end{equation}
The layer-wise \emph{alignment score} for $\text{ICD}_{i,p}$ at the $i$-th decoder layer is defined as:
\begin{equation}
s_{\text{align}}^{(i,p)}(l)=
      \operatorname{IoU}(\mathcal{R}_{i,p,l},\mathcal{B}_{i,p}),
\end{equation}
where $\operatorname{IoU}(\cdot,\cdot)$ denotes the computation of Intersection over Union (IoU) and $\mathcal{R}_{i,p,l}$ is the image region composed of $R_{i,p,l}$. We compute the alignment scores for every case in $\mathcal{G}_{\text{correct}}$ and $\mathcal{G}_{\text{wrong}}$. A higher $s_{\text{align}}$ indicates that the LVLM focuses more on regions within each image that are semantically aligned with the paired text during multimodal ICL.

\paragraph{ICD Contribution Perturbation.}
To study whether LVLMs can correctly locate the key ICD within the input sequence via self-attention, we sample $10,000$ additional sequences from $\mathcal{S}$. For each sequence, we keep one randomly chosen ICD unchanged, denoted $\text{ICD}^{\dagger}$, and replace the other two ICDs with samples from the HatefulMemes dataset. This setup ensures that every sequence $s_i$ contains one ICD that is semantically closest to the query sample, whereas the remaining two ICDs are almost irrelevant. 
For every sequence, three variants are generated by placing $\text{ICD}^\dagger$ in the first, second, or third position, respectively, yielding 30,000 perturbed sequences.  
Using the same correct/wrong split procedure described above, we obtain new $\mathcal{G}'_{\text{correct}}$ and $\mathcal{G'}_{\text{wrong}}$ sets, each containing 5,000 cases.

\paragraph{Saliency-Based Contribution Score $s_{\text{contrib}}$.}
To quantify the token-level information flow we compute a \emph{saliency score matrix} $I_{i}\in\mathbb{R}^{S_i^{\text{total}}\times S_i^{\text{total}}}$ at each $l$-th decoder layer:
\begin{equation}
I_{i,l} \;=\;\bigl\lvert\,A_{i,l} \;\odot\; \tfrac{\partial\mathcal{L}(s_i)}{\partial A_{i,l}}\bigr\rvert,
\end{equation}
where $A_{i,l}$ denotes the post-softmax attention matrix, $\mathcal{L}(s_i)$ is the loss function for input $s_i$, $\odot$ denotes element-wise (Hadamard) product, and the absolute value is applied element-wise. $I_{i,l}(m, n)$ reflects the significance of the information flow from the $n$-th token to the $m$-th token at the $l$-th layer.

Through perturbation, we place the key ICD at position $p\!\in\!\{1,2,3\}$ within a sequence $s_{i}$. Let $\mathbf{V}^{\dagger}_{i,p}$ denote the index set of visual tokens from that position,  
$\mathbf{V}_{i}$ denote the union index set of visual tokens from all three ICDs and the query sample, and $\mathbf{S}_i^{\text{ans}}$ denote the index set of the generated answer tokens.
The layer-wise \emph{contribution score} for position $p$ is defined as  
\begin{equation}
s^{(i,p)}_{\text{contrib}}(l) \;=\;
\frac{\displaystyle\sum_{v\in\mathbf{V}^{\dagger}_{i,p}}\sum_{t\in \mathbf{S}_i^{\text{ans}}} I_{i,l}(v,t)}{\displaystyle\sum_{v'\in\mathbf{V}_i}\sum_{t\in \mathbf{S}_i^{\text{ans}}} I_{i,l}(v',t)} ,
\end{equation}
i.e.\ the fraction of answer-directed saliency that can be traced back to the visual tokens of the key ICD in position $p$. A higher $s_{\text{contrib}}$ indicates that a larger proportion of the answer-related signal is attributable to the key ICD.
\subsection{3 $\quad$ Experiments}
\subsubsection{\textbf{3.1 Setup Details}}
\paragraph{Benchmarks.} In this work, to verify that CAMA can be broadly applied to various multimodal ICL tasks, we follow common evaluation protocols and include several more challenging benchmarks, using a total of eleven benchmarks:
\begin{itemize}
    \item \textbf{VQAv2}: VQAv2 contains 443,757 samples in the training set and 214,354 in the validation set. It is a classic VQA benchmark that tests a model’s ability to understand both the image and the question across diverse real-world scenarios. The images are sourced from the MSCOCO dataset, and the evaluation metric is Accuracy.
    \item \textbf{VizWiz}: VizWiz contains 20,523 samples in the training set and 4,319 in the validation set. It presents a more challenging setting with lower-quality images, along with many unanswerable cases, pushing models to handle uncertainty based on the format learned from ICDs. Its evaluation metric is Accuracy.
    \item \textbf{OK-VQA}: OK-VQA contains 9,055 samples in the training set and 5,000 in the validation set. It requires the model to incorporate external knowledge beyond the image content and the context to generate correct answers. The evaluation metric is Accuracy.
    \item \textbf{GQA}: GQA contains 943k samples in the training set and 132k in the validation set. This benchmark requires the model to perform multi-step, compositional reasoning when answering VQA questions, rather than relying on answer priors. The evaluation metric is Accuracy.
    \item \textbf{TextVQA}: TextVQA contains 34,602 samples in the training set and 5,000 in the validation set. Questions deliberately reference scene text (``What does the billboard say?"), forcing models to integrate OCR output with visual and linguistic cues. The evaluation metric is Accuracy.
    \item \textbf{CLEVR}: CLEVR Count Induction (CLEVR), a subset from VL-ICL bench, contains 800 samples in the training set and 200 in the validation set. Each ICD image is paired with an obscure query formatted as an attribute–number pair that identifies specific objects based on four attributes: size, shape, color, or material. We reformat them like ``(Q: Red. A: 2.)". The task requires models to perform challenging reasoning, first recognizing the task pattern, then counting the correct number of objects. The evaluation metric is Accuracy.
    \item \textbf{MMStar}: MMStar consists of 1,500 hand-picked, vision-indispensable samples balanced across 6 core skills and 18 fine-grained axes. Every item was filtered from 22,000 candidates to eliminate data leakage and questions answerable by language alone, so models cannot ``cheat" with memorized knowledge. Since it does not provide a predefined training and validation split, we divide the data using a 7:3 ratio. The evaluation metric is Accuracy, and the reported results are weighted averages across its six subtasks.
    \item \textbf{Flickr30k} \cite{flickr}: Flickr30k contains 29,783 samples in the training set and 1,000 in the validation set. It consists of images showing everyday activities, each paired with multiple human-written captions that provide concise descriptions of the scenes. The evaluation metric is CIDEr.
    \item \textbf{MSCOCO} \cite{mscoco}: MSCOCO contains 82,783 samples in the training set and 40,504 in the validation set. It is a widely used benchmark featuring a diverse range of images with detailed and richly descriptive captions, ideal for evaluating a model’s overall perception and description capabilities. The evaluation metric is CIDEr.
    \item \textbf{Hatefulmemes} \cite{hateful}: HatefulMemes contains 8,500 samples in the training set and 500 in the validation set. It is designed to reflect real-world challenges found in internet multimodal data. The task requires the model to jointly interpret both the image and the overlaid text to detect instances of hate speech. The evaluation metric is ROC-AUC.
    \item \textbf{L-I-VST} \cite{llava}: L-I-VST refers to the Visual StoryTelling subset of LLaVA-Interleave Bench, containing 400 samples. Each sample is an interleaved image-text sequence, where the model is required to generate a new sentence for a given image based on preceding image-sentence pairs, forming a coherent story.
\end{itemize} 

\paragraph{Models.} In this work, we evaluate CAMA on four LVLMs with diverse LLM backbones, training data, and numbers of tokens per image.
\begin{itemize}
    \item \textbf{LLaVA-NeXT-7B}: LLaVA-NeXT-7B uses CLIP-L/336 as the vision encoder and Mistral-7B as the LLM backbone. It is instruction-tuned on high-quality visual-instruction pairs that emphasize OCR and logical reasoning while reusing the efficient LLaVA-1.5 training recipe. It converts each image into up to 2,880 tokens for high-resolution inputs. Considering its context length, we set the number of tokens per image to 576.
    \item \textbf{Idefics2-8B}: Idefics2-8B uses SigLIP-SO400M as the vision encoder and Mistral-7B as the LLM backbone. Its training happens in two stages: (1) pre-training on OBELICS at 384$^2$ resolution, and (2) native-resolution fine-tuning ($\le$ 980 px) that adds OCR-heavy corpora such as PDFA, Rendered-Text and IDL, followed by instruction tuning on the 50-dataset Cauldron mix plus nine text-only sets. It converts each image into 64 tokens.
    \item \textbf{InternVL2.5-8B}: InternVL2.5-8B uses InternViT-300M as the vision encoder and InternLM 2.5-7B chat as the LLM backbone. It uses a three-stage curriculum (MLP warm-up → optional ViT incremental learning → full instruction tuning) combined with dynamic high-resolution tiling, JPEG-robust augmentation, and a stringent data-filtering pipeline for training. It can adjust the number of tokens per image based on input resolution. We set this number to 900 for consistency.
    \item \textbf{Qwen2.5VL-7B}: Qwen2.5VL-7B uses a window-attention ViT that supports dynamic spatial/temporal sampling and multi-resolution RoPE as the vision encoder and Qwen 2.5-7B as the LLM backbone. It is pretrained on 4.1 T text tokens plus large-scale image + video data, then instruction-tuned for chat-style vision-language tasks. It can also adjust the number of tokens per image based on input resolution. We set this number to 1156 for consistency. 
\end{itemize} 

\paragraph{Baselines.} Given that no existing method enhances multimodal ICL by modulating the intermediate attention logits of LVLMs and we are the first to do so, we select the following baselines for comparison:
\begin{itemize}
    \item \textbf{Vanilla}: Vanilla denotes directly feeding the LVLM with in-context sequences of the form $(I_1,T_1,...,I_n,T_n,Query Sample)$ to perform ICL.
    \item \textbf{+Inst}: Instruction augmented (+Inst) prepends to the vanilla sequence a sentence that explicitly instructs the model to carry out ICL: ``First, study the examples we provide. Then utilize what you have learned to answer the new question." Notably, this sentence achieves the best overall performance among the three styles we tested and is therefore adopted in our main experiments. The other two variants are ``Please make full use of the provided examples to solve a new image-related question" and ``Here are several example image-text pairs, along with a new image and question. Please perform in-context learning." \cite{h5}
    \item \textbf{CD}: Contrastive decoding (CD) constructs a distorted variant targeted to a selected part of the input, separately feeds the original and distorted sequences into the model, and collects their final output logits. By calibrating the vanilla logits with those from the distorted input, we obtain new logits whose distribution implicitly ensures that the model attends to the selected part. In LVLMs, CD is typically employed to encourage the model to rely more on visual evidence and thus mitigate hallucination or language prior \cite{h1,h2}. A common practice is to either add noise to the input image or directly replace it with a blank image. CD can likewise enhance multimodal ICL. We observe that replacing the images with blank images yields a larger improvement than adding noise. After calibration, the resulting logits accentuate the influence of the ICDs.
    Formally, CD is presented by:
    \begin{equation}
        (1 + \alpha) \, \text{logits}_{\theta}(y_t \mid y_{<t},X) 
- \alpha \, \text{logits}_{\theta}(y_t \mid y_{<t}, \hat{X}),
    \end{equation}
    where $\theta$ denotes the LVLM, $y_t$ denotes the model output at time step $t$, and $X$ and $\hat{X}$ denote the vanilla sequence and the distorted sequence, respectively. $\alpha$ is a user-defined value. For all VQA benchmarks, we set $\alpha = 0.4$. For other benchmarks, we set $\alpha = 0.6$ \cite{h3,h4}.
    \item \textbf{VE}: Visual enhancement (VE) is an effective strategy for improving LVLM performance that requires only changes in the input images. It works by manually highlighting the visual features that the user considers important, thereby increasing the model’s awareness of them. For each image in the ICDs of a vanilla sequence, we draw red bounding boxes around the regions mentioned by the accompanying text and feed the annotated sequence to the model. An image can contain multiple bounding boxes. Every annotation undergoes two additional human checks to ensure that the highlighted visual cues accurately reflect the textual semantics.
    \item \textbf{SoFA}: Soft Attention (SoFA) is a training-free and plug-and-play module for LVLMs that alleviates position bias by reshaping inter-image attention. Specifically, rather than employing purely causal (unidirectional) attention among visual tokens, SoFA defines a soft attention mask to add its bidirectional counterpart:
    \begin{equation}
    \mathbf{M}_{\mathrm{soft}} = (1 - \sigma)\,\mathbf{1}_{\mathrm{causal}}
                        + \sigma\,\mathbf{1}_{\mathrm{bidirectional}},
    \end{equation}
    and computes the resulting attention output as
    \begin{equation}
    \mathbf{H}_{\mathrm{soft}}
    = \mathrm{Softmax}\!\Bigl(\frac{QK^\top}{\sqrt{d}}\Bigr)
      \;\odot\;
      \mathbf{M}_{\mathrm{soft}}\,V,
    \end{equation}
    where \(\sigma\in[0,1]\) controls the interpolation degree between causal and bidirectional masks and is selected via a small validation set. The interpolation is applied every two layers to preserve the model’s original training dynamics. By smoothing positional cues and reinforcing early-image interactions while preserving text autoregression, SoFA achieves balanced reasoning over all image positions and significantly reduces prediction inconsistency. It has also been found to improve the performance of multimodal ICL.
\end{itemize}

\subsubsection{\textbf{3.2 Impact of Position Factor under Varying ICD Shot Counts}\\}

\begin{figure}[t] 
  \centering                 
  \includegraphics[width=\columnwidth]{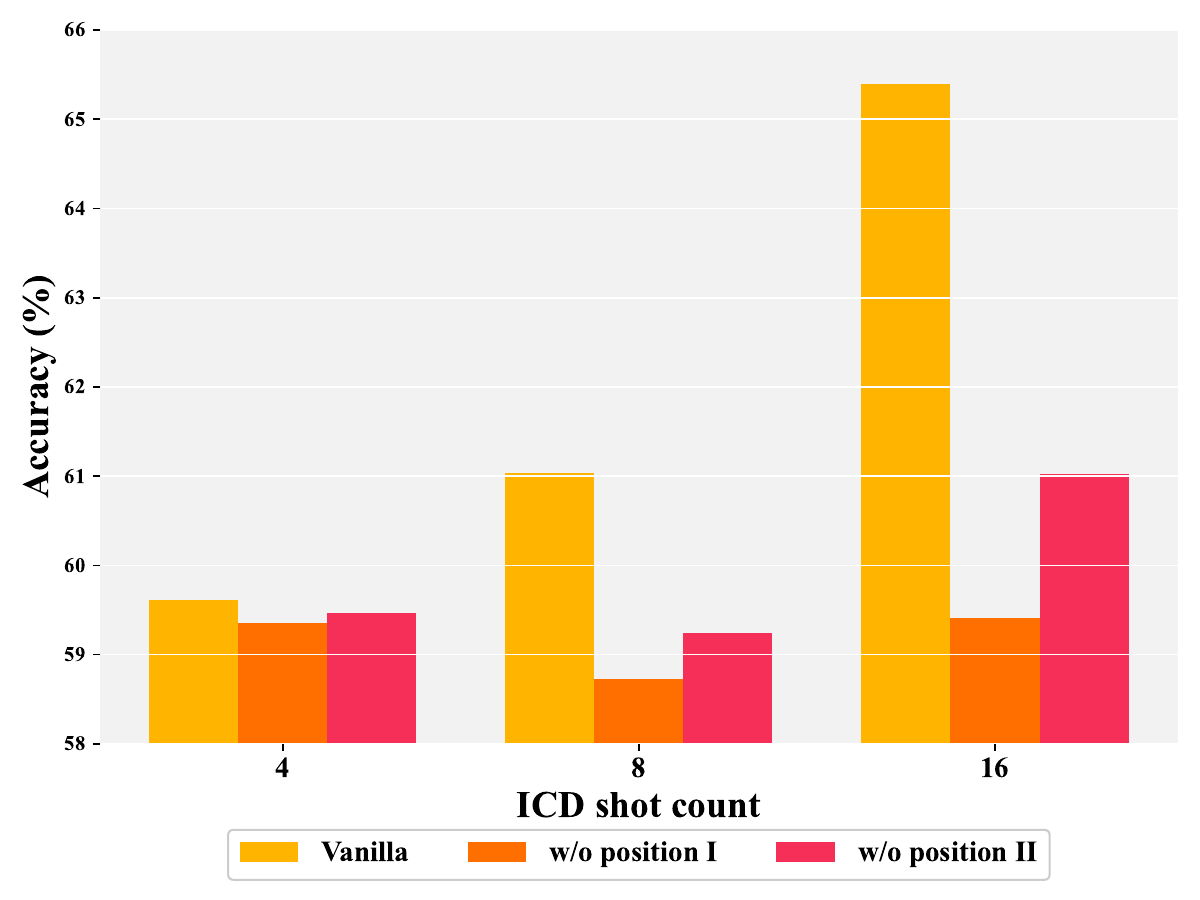}
  \caption{Average performance comparison of CAMA (Vanilla vs. w/o position-decay factors) under different shot counts.} 
  \label{shot2}          
\end{figure}

Figure~\ref{shot2} reports the performance drop that occurs when the position-decay factors are removed from Stage~I or Stage~II at shot counts of $4$, $8$, and $16$.  As the number of shots increases, the benefit contributed by the position factor grows steadily, which corroborates \textbf{\textit{Finding~3}}.  With more shots in a sequence, earlier ICDs suffer greater attention deficits. The position factor compensates for this imbalance and thus becomes crucial when extending CAMA to long-context and many-shot scenarios.

\subsubsection{\textbf{3.3 Impact of the Selection of Anchor Tokens}\\}

\begin{table}[t]
\centering
\begingroup
\setlength{\tabcolsep}{1.6pt}
\footnotesize
\begin{tabular}{cccccccc}
\toprule
\textbf{Method} & \textbf{VQAv2} & \textbf{VizWiz} & \textbf{OK-VQA} & \textbf{GQA} & \textbf{TextVQA}\\ 
\midrule
$\mathbf{S}^{Q}_{i}[0]$ & 68.12 & 50.69 & 60.56 & 68.60 & 77.53 \\
$\mathbf{S}^{I}_{i}[-2]$ & 67.34 & 49.83 & 60.03 & 67.84 & 76.91 \\
$\mathbf{S}^{I}_{i}[-1]$ & 67.15 & 49.37 & 59.12 & 67.75 & 76.58\\
\midrule
$\mathbf{S}^{A}_{i}[0]$ & 68.12 & 50.69 & 60.56 & 68.60 & 77.53 \\
$\mathbf{S}^{Q}_{i}[-1]$ & 67.93 & 50.92 & 60.27 & 68.35 & 77.16 \\
$\mathbf{S}^{Q}_{i}[-2]$ & 67.87 & 50.42 & 59.84 & 68.39 & 76.95 \\
$\text{Random}(\mathbf{S}^{Q})$ & 67.21 & 49.92 & 59.15 & 67.78 & 76.71 \\
\midrule
$\mathbf{S}^{A}_{i}[-1]$ & 68.12 & 50.69 & 60.56 & 68.60 & 77.53 \\
$\mathbf{S}^{Q}_{i}[-2]$ & 68.04 & 50.48 & 60.71 & 68.41 & 76.82 \\
$\text{Random}(\mathbf{S}^{A})$ & 67.32 & 50.13 & 60.24& 68.34 & 76.29 \\
\bottomrule
\end{tabular}
\caption{Average performance of CAMA with different three-anchor-token combinations. The first row in each block corresponds to the original anchor-token selection. $\text{Random}(\cdot)$ denotes a token randomly chosen from the sequence $\cdot$.}
\label{tab:appat}
\endgroup
\end{table}
In computing the dynamic attention increments, the choice of the three anchor tokens is critical. For efficiency and owing to the semantic accumulation that emerges in the middle layers, we select $\mathbf{S}^{Q}_{i}[0]$, $\mathbf{S}^{A}_{i}[0]$, and $\mathbf{S}^{A}_{i}[-1]$ as the anchor tokens. We then vary each of these anchor tokens to investigate their influence on CAMA performance, and the results are presented in Table \ref{tab:appat}. The findings show that semantic accumulation indeed facilitates the calculation of dynamic attention increments, allowing the last tokens to summarize the semantics of the preceding tokens. In contrast, the final tokens of an image are less representative and cannot serve as suitable anchor tokens, which further indicates that attention deficits also exist within image tokens and that their semantics often drift toward the subsequent text.
\subsubsection{\textbf{3.4 Additional Qualitative Visualizations}\\}
\begin{figure}[t] 
  \centering                 
  \includegraphics[width=\columnwidth]{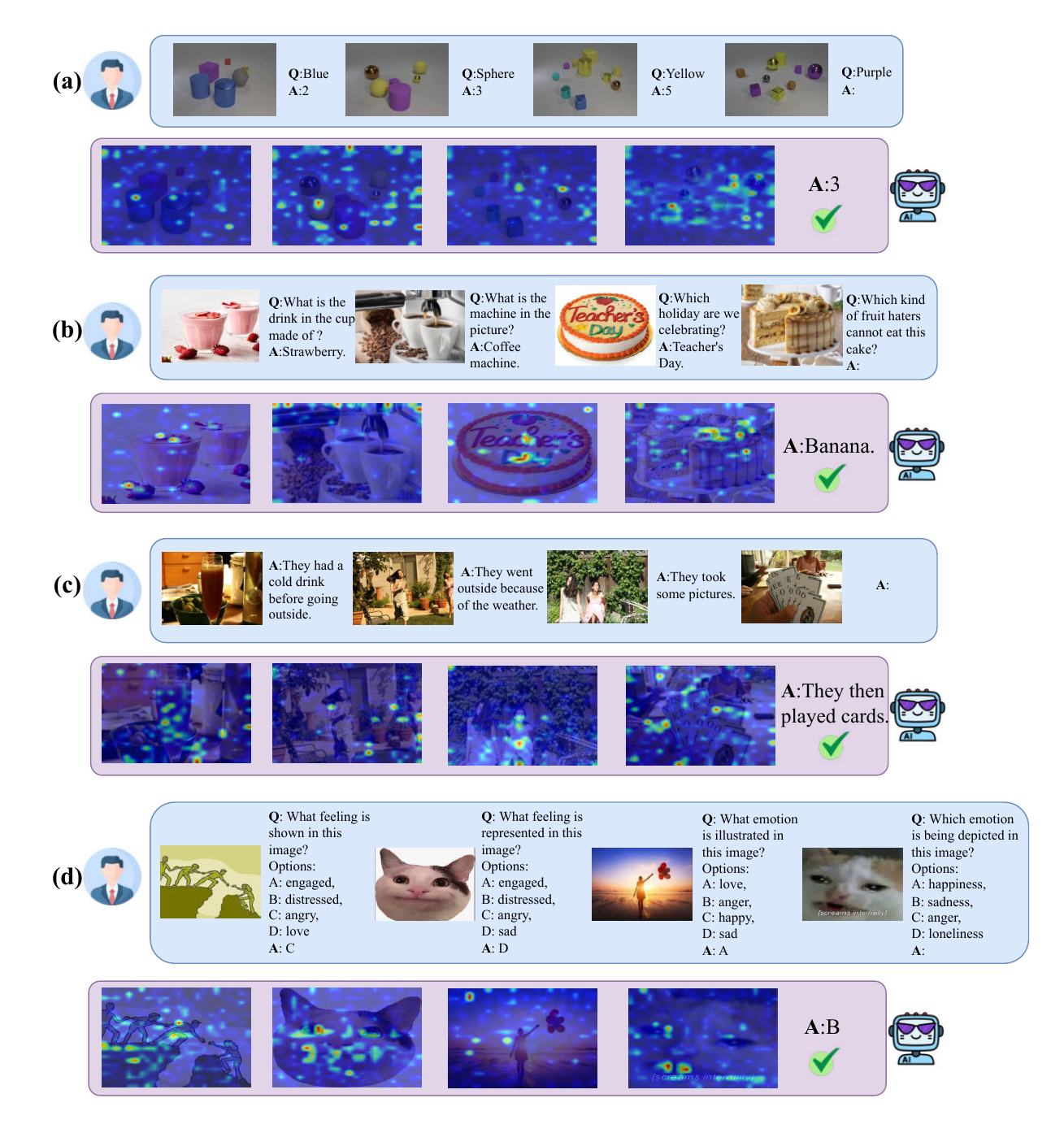}
  \caption{Qualitative visualizations of CAMA on four cases.} 
  \label{case1}          
\end{figure}
In Figure \ref{case1}, we present the ICL enhancement results of CAMA for four input sequences by visualizing the image attention in the 19th decoder layer of LLaVA-NeXT-7B. In (a), the sequence comes from CLEVR. This benchmark poses a challenge for the model to identify the task mapping in ICL because of its subtle textual descriptions. Through dynamic attention increments, CAMA highlights the key visual features that align with the text semantics, helping the model infer the correct task mapping. As shown in the figure, the model accurately locates the objects whose attributes match those in the question and successfully performs the counting task. In (b), the example illustrates a typical VQA scenario. In this sequence, although the third ICD and the query sample both depict cakes as the main subject, the first ICD is actually more crucial for solving the query sample, as both focus on identifying food ingredients. CAMA effectively prevents the first ICD from being overlooked due to position bias, guiding the model’s attention toward the connection between ``drink" and ``strawberry" in the first ICD. This association helps the model identify the relationship between ``cake" and ``banana" in the query sample, enabling it to provide the correct answer. In (c), the example is from L-I-VST, which requires the model to complete an interleaved story. While reinforcing the alignment between each image–text pair, CAMA also guides the model to focus on consistent visual cues between earlier images and the query sample image, such as ``a girl wearing pink clothes," thereby recognizing the narrative as a coherent story. This enables the model to generate text that follows the format established in the preceding context. In (d), the sequence comes from MMStar, which presents tasks in a multiple-choice format and requires the LVLM to recognize the emotion in an image. CAMA identifies the second ICD as the most helpful for answering the query sample, since both images are cat memes. The perception of the second image is thus enhanced, allowing it to fully capture the connection between the cat’s expression and its ground-truth option ``sadness." When interpreting the query sample image, the model focuses more on the cat’s expression rather than the text in the image, demonstrating CAMA’s ability to deeply promote effective multimodal ICL. However, in our experiments we also find that CAMA struggles when the in-context sequence contains too much irrelevant information, such as cases where seven out of eight or all ICDs are unrelated to the query sample or provide misleading content. In these situations, the difficulty of appropriately allocating attention in Stage II prevents the model’s behavior from being steered in the desired direction. These failure cases highlight that improving the quality of sequence configuration remains important.
\subsubsection{\textbf{3.5 Efficiency Analysis}\\}
\begin{table}[t]
  \centering
  \begin{tabular}{c|cc}
    \toprule
    \textbf{Method} & \textbf{Inference Latency} & \textbf{Accuracy} (\%) \\
    \midrule
    Vanilla & 4.70s & 58.07 \\
    CD & 9.42s & 57.95 \\
    SoFA & 4.72s& 59.15 \\
    CAMA & 4.94s& 61.03 \\
    \bottomrule
  \end{tabular}
  \caption{Comparison of average inference latency and Accuracy of four methods in the 8-shot setting.}
  \label{tab:latency}
\end{table}
As shown in Table \ref{tab:latency}, CAMA introduces only a small increase in inference latency compared with the vanilla model and SoFA, while delivering substantial performance gains. Compared with CD, which requires two forward passes, CAMA also achieves a clear efficiency advantage. When the number of ICDs is smaller, the added inference latency of CAMA further reduces. Therefore, although CAMA incurs some additional cost, its cost-effectiveness in practical applications remains superior.

\end{document}

%% file: maintab.tex
\begin{table*}[t]
\centering
\footnotesize
\begin{tabular}{ccccccccc|c}
\toprule
\textbf{LVLM} & \textbf{Method} & \textbf{VQAv2} & \textbf{VizWiz} & \textbf{OK-VQA} & \textbf{GQA} & \textbf{TextVQA} & \textbf{CLEVR} & \textbf{MMStar} &\textbf{Avg.}\\ 
\hline
\multirow{8}{*}{LLaVA-NeXT} 
  & Vanilla      & 61.86 & 37.64 & 57.63 & 55.38 & 61.93 & 16.50 & 44.72 & 47.95 \\
  & +Inst        & 61.69 & 38.52 & 58.21 & 55.70 & 61.74 & 17.46 & 43.95 & 48.18 \\
  & CD           & 61.79 & 37.70 & 57.48 & 55.46 & 62.07 & 17.18 & 41.59 & 47.61 \\
  & VE           & 61.94 & 37.58 & 57.97 & 55.29 & 61.78 & 18.16 & 44.92 & 48.23 \\
  & SoFA         & 63.21 & 38.09 & 58.14 & 57.42 & 62.28 & 14.29 & 45.71 & 48.45 \\
  \rowcolor{lightgray}
  \cellcolor{white}&CAMA         & 64.46 & 39.87 & 59.94 & 58.60 & 63.40 & 18.67 & 47.16 & 50.30 \\
  \rowcolor{lightgray}
  \cellcolor{white}& CAMA(+Inst) & \underline{64.89} & \underline{40.23} & \underline{60.27} & \underline{58.71} & \underline{63.61} & \underline{21.28} & \underline{47.42} & \underline{50.92} \\
  \rowcolor{lightgray}
  \cellcolor{white}& CAMA(VE)    & \textbf{65.24} & \textbf{40.67} & \textbf{60.58} & \textbf{59.04} & \textbf{64.07} & \textbf{23.17} & \textbf{47.71} & \textbf{51.50} \\
\hline
\multirow{8}{*}{\cellcolor{white}Idefics2} 
  & Vanilla      & 57.32 & 38.46 & 43.60 & 57.49 & 70.02 & 34.61 & 42.65 & 49.16 \\
  & +Inst        & 57.61 & 38.25 & 43.75 & 57.38 & 70.30 & 35.48 & 42.51 & 49.21 \\
  & CD           & 56.83 & 38.19 & 43.47 & 57.30 & 68.87 & 33.70 & 41.49 & 48.55 \\
  & VE           & 57.28 & 38.89 & 44.26 & 57.98 & 71.18 & 36.41 & 42.93 & 49.85 \\
  & SoFA         & 59.04 & 38.95 & 46.12 & 57.75 & 72.31 & 34.44 & 43.29 & 50.27 \\
  \rowcolor{lightgray}
  \cellcolor{white}& CAMA         & 60.53 & 39.90 & 47.23 & 59.79 & 74.38 & 36.52 & \underline{44.86} & 51.89 \\
  \rowcolor{lightgray}
  \cellcolor{white}& CAMA(+Inst) & \underline{60.74} & \textbf{40.37} & \underline{47.69} & \underline{60.00} & \textbf{76.21} & \underline{38.95} & 44.59 & \underline{52.40} \\
  \rowcolor{lightgray}
  \cellcolor{white}& CAMA(VE)    & \textbf{60.82} & \underline{40.16} & \textbf{47.80} & \textbf{60.08} & \underline{75.72} & \textbf{39.58} & \textbf{45.27} & \textbf{52.78} \\
\hline
\multirow{8}{*}{\cellcolor{white}InternVL2.5} 
  & Vanilla      & 69.58 & 58.27 & 62.32 & 67.21 & 80.29 & 56.91 & 62.70 & 65.33 \\
  & +Inst        & 69.89 & 58.92 & 62.18 & 67.53 & 81.10 & 57.34 & 62.38 & 65.53 \\
  & CD           & 69.81 & 58.79 & 63.01 & 67.54 & 80.17 & 57.11 & 62.56 & 65.57 \\
  & VE           & 69.80 & 58.96 & 63.14 & 67.28 & 80.39 & 59.53 & 62.95 & 66.00 \\
  & SoFA         & 70.85 & 59.62 & 62.48 & 68.30 & 82.75 & 59.12 & 62.70 & 66.55 \\
  \rowcolor{lightgray}
  \cellcolor{white}&CAMA         & 72.54 & 62.15 & \underline{66.27} & 70.68 & 85.19 & 61.45 & \textbf{64.27} & 68.94 \\
  \rowcolor{lightgray}
  \cellcolor{white}& CAMA(+Inst) & \underline{72.61} & \underline{62.48} & 65.58 & \underline{70.93} & \underline{85.46} & \underline{64.67} & 63.37 & \underline{68.97} \\
  \rowcolor{lightgray}
  \cellcolor{white}& CAMA(VE)    & \textbf{72.85} & \textbf{64.57} & \textbf{66.73} & \textbf{71.30} & \textbf{85.66} & \textbf{64.90} & \underline{63.79} & \textbf{69.97} \\
\hline
\multirow{8}{*}{\cellcolor{white}Qwen2.5VL} 
  & Vanilla      & 71.94 & 57.39 & 65.70 & 82.31 & 83.61 & 62.83 & 65.18 & 69.85 \\
  & +Inst        & 72.43 & 57.80 & 66.18 & 83.57 & 83.72 & 63.65 & 66.03 & 70.43 \\
  & CD           & 72.31 & 57.35 & 66.07 & 82.49 & 83.76 & 62.72 & 65.63 & 70.05 \\
  & VE           & 72.69 & 59.30 & 66.41 & 84.55 & 83.92 & 63.26 & 66.47 & 70.94 \\
  & SoFA         & 72.31 & 59.06 & 67.28 & 84.52 & 84.34 & 64.85 & 66.87 & 71.32 \\
  \rowcolor{lightgray}
  \cellcolor{white}&CAMA         & \underline{74.96} & 60.83 & 68.80 & \underline{85.32} & \underline{87.14} & 66.15 & \textbf{67.78} & 73.00 \\
  \rowcolor{lightgray}
  \cellcolor{white}& CAMA(+Inst) & 74.79 & \underline{61.28} & \underline{69.21} & 85.26 & 86.87 & \textbf{67.02} & 67.39 & \underline{73.12} \\
  \rowcolor{lightgray}
  \cellcolor{white}& CAMA(VE)    & \textbf{75.22} & \textbf{62.37} & \textbf{69.74} & \textbf{85.69} & \textbf{87.89} & \underline{66.81} & \underline{67.70} & \textbf{73.63} \\
\bottomrule
\end{tabular}
\caption{Accuracy of 8-shot ICL on seven VQA benchmarks for four LVLMs under different enhancement methods. The highest value is highlighted in \textbf{bold}, and the second highest is \underline{underlined}. The cells shaded in light gray present the results of CAMA.}
\label{tab:main}
\end{table*}